\documentclass[10pt,letterpaper]{article}

\usepackage{cogsci}

\cogscifinalcopy 

\usepackage{cmap}
\usepackage[T1]{fontenc}
\usepackage[american]{babel}
\usepackage{csquotes}
\usepackage{newtxtext,newtxmath}  
\usepackage{graphicx}
\usepackage[
  backend=biber,
  style=apa,
  natbib=true,
  annotation=false,
]{biblatex}
\usepackage{float} 

\usepackage[hidelinks]{hyperref}

\usepackage{xcolor}
\usepackage{hyperref}

\definecolor{linkblue}{RGB}{30,80,130}

\hypersetup{
  colorlinks=true,
  urlcolor=linkblue,
  linkcolor=black,
  citecolor=black
}

\title{Leveraging Speech to Identify Signatures of Insight and Transfer in Problem Solving}

\author[1]{\mbox{Linas Nasvytis (linasmn@stanford.edu)}}
\author[1,2,3]{\mbox{Judith E. Fan}}
\affil[1]{Department of Psychology, Stanford University}
\affil[2]{Graduate School of Education, Stanford University}
\affil[3]{Department of Computer Science, Stanford University}

\begin{document}

\maketitle

\begin{abstract}
Many problems seem to require a flash of insight to solve.
What form do these sudden insights take, and what impact do they have on how people approach similar problems in the future?
In this work, we prompted participants ($N=189$) to think aloud as they attempted to solve a sequence of five ``matchstick-arithmetic'' problems.
These problems either all relied on the same kind of non-obvious solution (Same group) or a different kind each time (Different group). Our first observation was that Same participants improved more rapidly than Different participants. We then leveraged techniques from natural language processing to analyze participants’ speech, and found that this accelerated improvement for Same participants was accompanied by changes in both how much they spoke and what they said. In particular, they were more likely to spontaneously label the kind of problem they were working on.
Taken together, these findings suggest that a hallmark of transferable insights is their accessibility for verbal report, even if the underlying precursors of insight remain difficult to articulate.

\textbf{Keywords:}
reasoning; language; strategies; learning; think-aloud protocol
\end{abstract}

\section{Introduction}

Many problems seem intractable until suddenly they are not.
Such moments of insight are often described as abrupt transitions, accompanied by a subjective ``Aha'' experience \citep{danek2017false, bowden2005new, kounios2014cognitive}.
Yet insight itself is not directly observable, and solving an ``insight problem'' does not guarantee that a solver experienced a discrete Aha moment.
Here we focus on just the observable behaviors that we treat as a proxy for these moments of insight. 
In particular, we investigate the distinguishing characteristics of the speech people produce as they are about to succeed in solving a problem for the first time, as well as what they say when they encounter similar problems in the future. 

A prominent account characterizes insight as a shift from search within an initial problem representation to search within an alternative representation, or ``problem space'' \citep{kaplan1990search}.
Such shifts may involve relaxing implicit constraints, decomposing familiar chunks, or discovering a more productive description of the problem state \citep{knoblich1999constraint, ollinger2008investigating}.
If successful solutions in such tasks reflect changes in the problem representation, the first successful attempt should do more than terminate the current search --- it may also change how solvers organize their later search: which candidate moves they consider, which features of the problem they attend to, and how readily they can apply the discovered structure to new problems.
These predictions raise the question of how these new approaches to search over the problem space become observable in subsequent reasoning.

Testing these predictions requires measures that capture problem solving as it unfolds, rather than only accuracy or response time.
Think-aloud protocols offer such a window by asking participants to verbalize their thoughts during a task \citep{ericsson1980verbal, ericsson2017protocol}.
However, these protocols have historically been difficult to scale because of the costs associated with transcribing speech and coding for the relevant features.
Recent advances in automated speech recognition and natural language processing have made it more feasible to analyze verbal behavior at multiple levels, including when and how much people talk, as well as what they are talking about, and what role each utterance seems to play in their reasoning process \citep{wurgaft2025scaling}.

Here, we use scalable analyses of think-aloud speech to study successful problem solving and transfer in matchstick-arithmetic problems, a classic insight paradigm in which solvers must transform an invalid equation by moving a single matchstick \citep{knoblich1999constraint, ollinger2006heuristics, ollinger2008investigating}.
In our experiment, participants solved a sequence of five invalid Roman numeral equations while thinking aloud.
We manipulated whether all five problems shared the same underlying transformation type (\emph{Same} condition) or each required a different transformation type (\emph{Different} condition).
This design distinguished two stages: the first successful attempt --- our operational marker of restructuring --- and subsequent trials, on which the discovered structure was reusable (\emph{Same}) or not (\emph{Different}).

This design enabled us to test two things: whether participants spoke differently on first-success trials than on earlier trials, across the multiple levels of speech we measured, and how participants continue to change how they speak after the first success when the discovered structure can be reused.
We found that on first-success trials in both groups, participants spoke more, talked about different things, and considered fewer candidate solutions.
After the first success, participants continued to change how they spoke in later trials only when the same problem type recurred in subsequent problems (\emph{Same} condition): they continued speaking more, kept shifting what they talked about, and increasingly named the problem type.
These findings suggest that successful restructuring does more than improve task performance: it changes how people talk about a problem.

\section{Methods}

\begin{figure*}[t]
  \centering
  \includegraphics[width=0.9\linewidth]{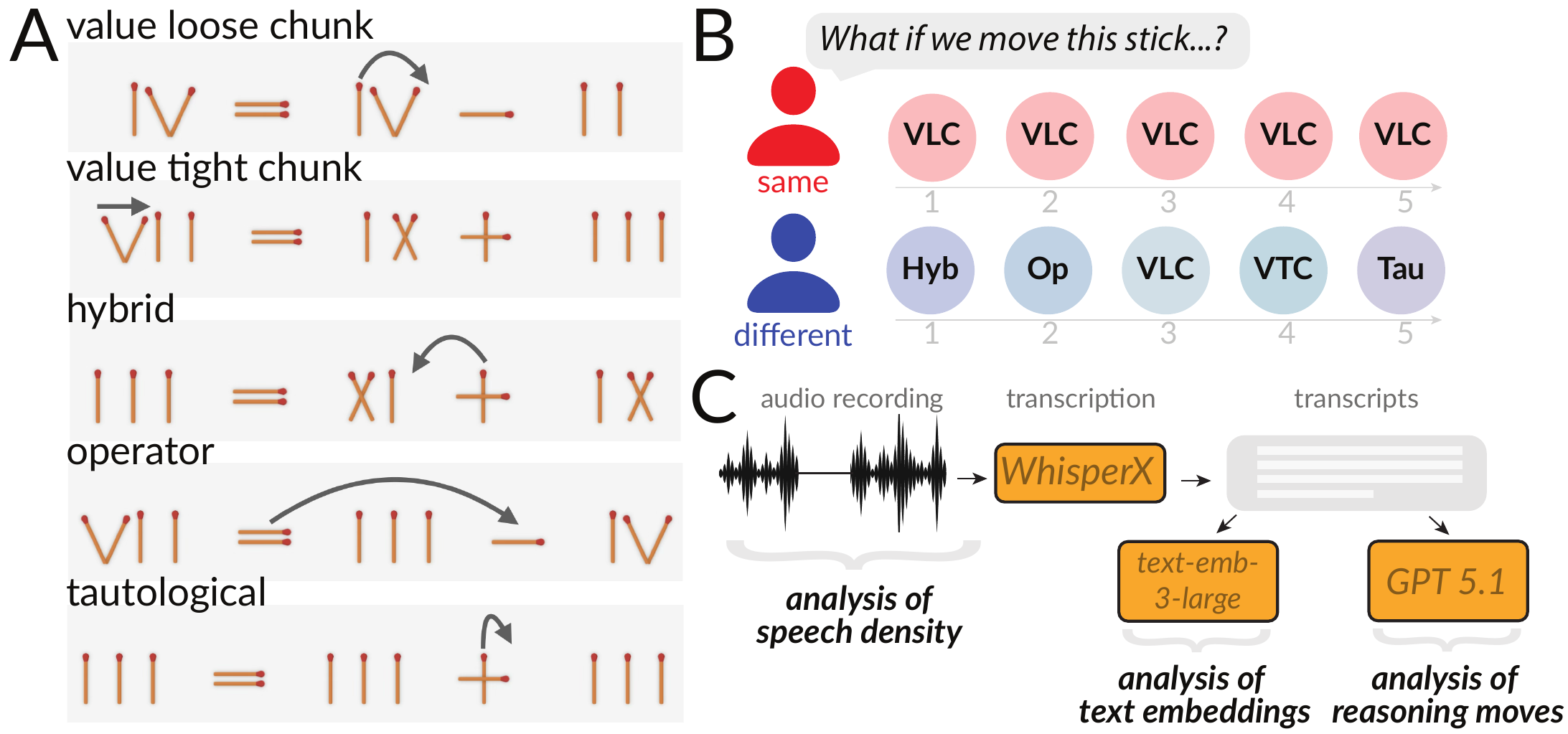}
  \caption{Overview of tasks, experimental design, and analysis pipeline. (A) Five matchstick arithmetic problem types, each defined by a distinct underlying transformation required for solution. (B) Experimental procedure in the Same and Different conditions across five trials. (C) Analysis pipeline linking audio recordings to speech, semantic, and reasoning analyses.}
  \label{fig:methods}
\end{figure*}

\subsection{Participants}
We recruited 202 fluent English speakers in the US via Prolific. Participants were compensated \$8.25 for a 30-minute task and gave informed consent. Preregistered exclusion criteria included failure to pass instructional practice, audio recording issues, and incomplete data. After exclusions, the final sample consisted of 189 participants (age: $M = 41.43$, $SD = 11.83$, range = 19--74; 103 female, 85 male, and 1 non-binary).

\subsection{Stimuli}
Participants solved five matchstick arithmetic problems while thinking aloud. Each trial presented an invalid Roman numeral equation constructed from matchsticks, which could be made valid by moving a single stick. These problems have been widely used to study insight problem solving, including representational change and constraint relaxation \citep{knoblich1999constraint, ollinger2006heuristics, ollinger2008investigating}. We created 25 novel problems grouped into five types (five per type), each defined by a distinct transformation (e.g., changing a numeral or an operator). Surface features varied across instances, but all followed the same visual format and length. All problems were solvable via one-stick moves; when multiple solutions existed, they involved the same underlying transformation. Each problem type is visualized in Figure~\ref{fig:methods}A. 

\subsection{Design \& Procedure}
Participants were randomly assigned to one of two between-subjects conditions (Figure~\ref{fig:methods}B). In the \emph{Same} condition, all five problems shared the same transformation type, allowing the same problem type to repeat across trials. In the \emph{Different} condition, each problem required a different transformation type, so the same problem type did not repeat across trials.

The task was completed online using a custom web-based interface. After providing informed consent, participants completed a microphone check and two instructional practice blocks assessing Roman numeral recognition and basic arithmetic. Both practice blocks had to be passed in order to proceed. Participants were then shown a worked example and introduced to the matchstick arithmetic task.

On each trial, participants viewed a static image of an equation and were instructed to think aloud while solving the problem. Audio was recorded continuously throughout each trial. Participants entered responses via an on-screen interface and received immediate binary feedback. Incorrect responses could be revised and resubmitted within the four-minute time limit. If the time limit expired without a correct solution, the trial ended without revealing the answer. An automated prompt reminded participants to verbalize if extended silence was detected. Trial order was randomized within condition. 

\subsection{Measures and Data Analysis}
We analyzed performance, speech, and language at the trial and utterance levels to characterize how problem solving and verbal behavior varied with success and transfer. Analyses focused on differences by trial correctness, phase (relative to the first success), and experimental condition. The analysis pipeline is summarized in Figure~\ref{fig:methods}C.

\subsubsection{Task performance}
For each trial, we recorded accuracy (whether a correct solution was produced within the time limit) and response time (time to correct submission). To examine phase-dependent effects, trials were labeled as \emph{pre-success}, \emph{first-success}, or \emph{post-success}, with post-success defined as correctly solved trials following a participant’s first correct solution. 

\subsubsection{Statistical models}
Unless otherwise noted, trial-level outcomes were analyzed with mixed-effects
models that included condition, problem type, and participant-level random effects. Accuracy was analyzed with logistic mixed-effects models. Response time
was log-transformed and analyzed only on correct trials, because incorrect
trials ended at the four-minute time limit. Speech density and speech rate were
analyzed with linear mixed-effects models. Models testing change across phases
included phase, condition, their interaction, problem type, and participant-level
random intercepts.
\subsubsection{Speech output}
Audio recordings were segmented into speech and silence using Silero voice activity detection, from which we computed speech density (the proportion of trial time with speech). Recordings were transcribed using WhisperX and split into time-aligned segments, which we refer to as \textit{utterances}. Speech rate was computed as the total number of words divided by the trial duration. Speech density was also computed across five equal-duration bins to capture within-trial dynamics. Two participants with at least one trial lacking any transcribed words were excluded from speech-based analyses (final speech sample: $N = 187$). Behavioral analyses retained the full sample.

\subsubsection{Semantic content}
Each utterance was embedded using the OpenAI \texttt{text-embedding-3-large}
model. We trained logistic regression classifiers to predict correctness, trial
phase, and condition from these embeddings. Cross-validation folds were constructed within-participant, so all utterances from a given participant appeared in either the training set or the test set, but not both. These classifiers tested whether what participants said varied by correctness, phase, or condition.

\subsubsection{Reasoning moves}
To track what role each utterance played in problem solving, we used GPT-5.1 to annotate each utterance with one of seven mutually exclusive reasoning-move labels: \textbf{Proposal} (introducing a concrete candidate move or solution), \textbf{Evaluation} (computing outcomes or judging whether a candidate works), \textbf{Categorization} (recognizing the problem type or noting similarity to prior problems), \textbf{Restatement} (reading or paraphrasing the given equation or task constraints), \textbf{Affect} (emotional reactions such as frustration or excitement), \textbf{Filler} (non-contentful hesitations or interjections), and \textbf{Meta \& Other} (search regulation or strategic talk without a concrete candidate, including residual task-relevant comments not captured by other categories). The model was prompted with the utterance text, trial context, label definitions, and example annotations, and returned a single label for each utterance. We manually inspected a subset of annotations to verify that the labels matched the intended definitions.


\section{Results}
\subsection{Task Performance and Behavioral Evidence of Transfer}
\begin{figure}[t]
  \centering
  \includegraphics[width=\linewidth]{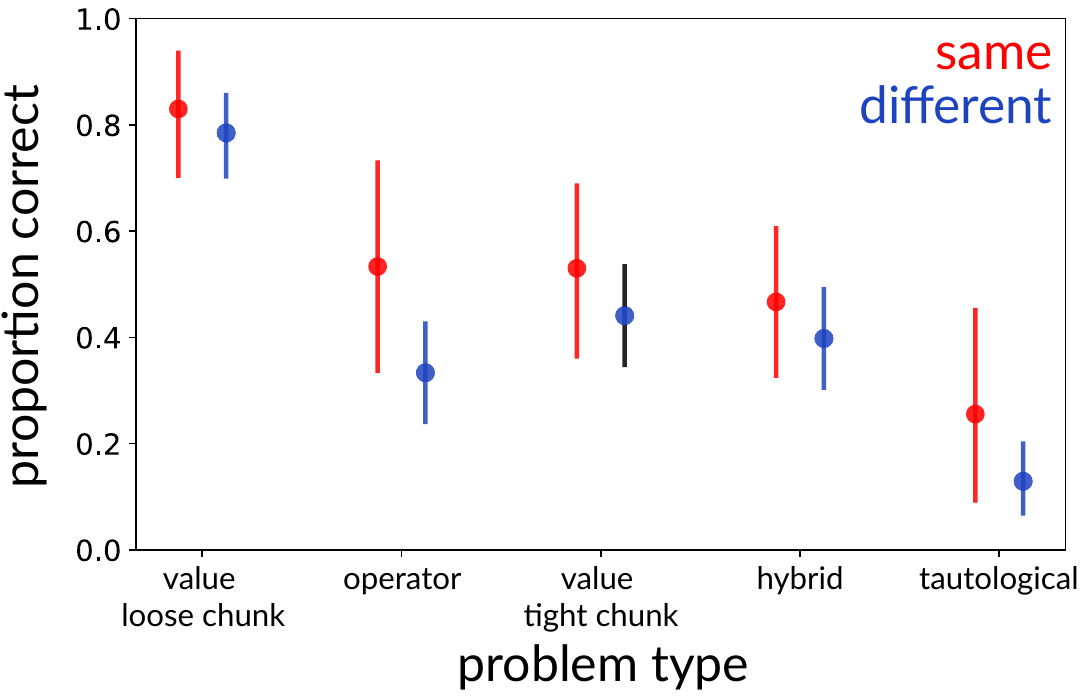}
    \caption{Accuracy by problem type, split by group (trial-level means; error bars indicate participant-bootstrap 95\% confidence intervals).}
  \label{fig:accuracy_by_type}
\end{figure}

\subsubsection{Problem difficulty varies across problem types}
Across all completed trials, participants solved 47.5\% of problems correctly (449 out of 945). Accuracy varied by problem type (Fig.~\ref{fig:accuracy_by_type}): \emph{value-loose-chunk} problems were easiest (80.7\%), while \emph{tautological} problems were most difficult (19.2\%). The remaining types (\emph{value-tight-chunk}, \emph{hybrid}, and \emph{operator}) showed intermediate accuracy.

These results confirm that the task spanned a wide range of difficulty, making it suitable for studying learning and transfer.

\subsubsection{Accuracy and speed improve more when structure repeats}
We next tested how performance changed across trials in each condition. We used
mixed-effects models to predict trial-level accuracy and response time from
trial number, condition, and their interaction, with random intercepts for
participants and random slopes for trial number.

Participants showed greater improvement across trials when the same problem
type repeated. In the fixed-effects predictions from the mixed-effects logistic
regression, accuracy in the Same condition rose from 0.32 on Trial~1 to 0.75 on
Trial~5, whereas accuracy in the Different condition remained between 0.32 and
0.39. The model supported this condition difference in trial-by-trial change
(trial~$\times$~group interaction: $\beta = -0.524$, $p < .001$). Participants
in the Different condition were also less accurate overall
($\beta_{\mathrm{group}} = -0.747$, $p = .031$).

A similar pattern emerged for response times on correct trials. Because unsolved trials always ended at the 240-second time limit, we compared response times on correct trials only. Participants solved correct trials faster when the same problem type repeated. In the mixed-effects model predicting log response time, the decrease across trials was smaller in the Different condition than in the Same condition (trial~$\times$~group interaction: $\beta = 0.079$, $p = .028$). Participants in the Different condition were also slower at the centered trial number ($\beta_{\mathrm{group}} = 0.362$, $p < .001$).

Together, these results show that accuracy and speed improve more when problem type is repeated across trials.

\subsubsection{Early success boosts later performance more when problem type is shared}


Finally, we asked how well participants performed \textit{after} their first success. We
analyzed participants who solved at least one problem by Trial~4, leaving at
least one later trial on which they could reuse the discovered structure. In the
Same condition, participants solved 83.7\% of subsequent problems correctly
($SE = 0.038$), compared with 44.0\% in the Different condition ($SE = 0.033$). They also solved later problems faster in the Same condition ($M = 89.9$~s, $SE = 8.25$) than in the Different condition ($M = 179.0$~s, $SE = 5.55$).



These data suggest that following their first success, participants solved subsequent problems more quickly and accurately when they were of the same type.

\subsection{Verbal Signatures of Successful Problem Solving}


Having shown that participants improved more after the first success when the
same problem type repeated, we next asked how their speech changed across the
same phases. We measured how much time participants spent speaking in each trial
(\textit{speech density}) and how many words they produced per second (\textit{speech rate}). These measures allowed us to ask how much participants talked as they moved
from unsuccessful trials, to the first success, to later trials.

\begin{figure}[t]
  \centering
  \includegraphics[width=\linewidth]{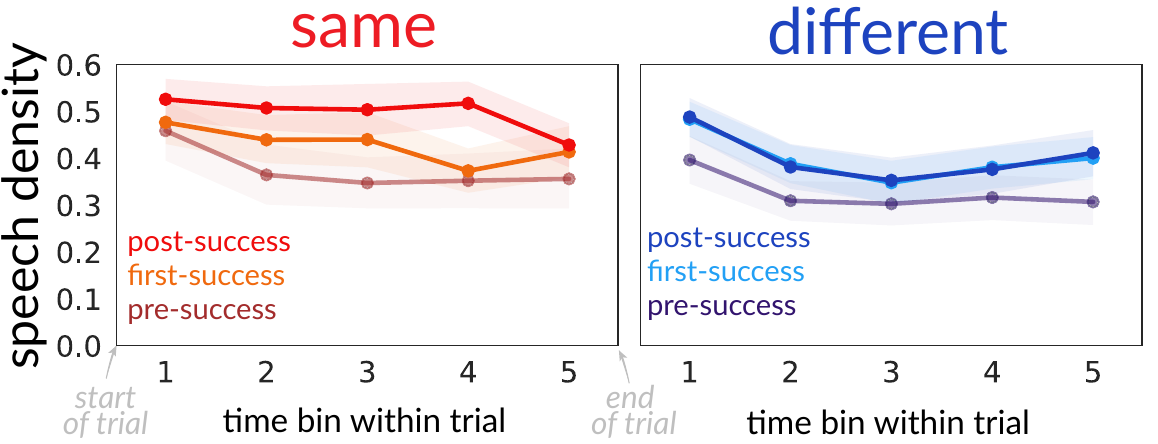}
  \caption{Speech density (speech–silence ratio) across the course of a trial, computed from five equal temporal segments and shown separately for each trial phase. Panels correspond to experimental groups (trial-level means; error bars indicate participant-bootstrap 95\% confidence intervals).}
  \label{fig:speech_density}
\end{figure}

\subsubsection{Participants speak more on correct trials}



We first tested how speech density differed between correct and incorrect trials. Across both conditions, participants spoke more on correct trials than on incorrect trials ($\Delta = 0.067$, $SE = 0.009$, $t = 7.71$, $p < .0001$). This correct--incorrect difference was smaller in the Different condition than in the Same condition (correctness-by-condition interaction: $\beta = -0.057$, $p = .0009$). Thus, participants spent more time speaking on trials they solved correctly, especially when the same problem type repeated.

\subsubsection{Participants speak more on the first successful trial than on the ones preceding it}


We next tested how speech density changed when participants solved their first 
problem correctly. We compared trials before the first success (\emph{pre-success}) with
the first-success trial. Participants spoke more at the first success in both
conditions. In the Same condition, speech density rose by 0.057 ($p = .0004$);
in the Different condition, it rose by 0.040 ($p = .0076$). The size of this
increase did not differ by condition. Thus, participants spoke more at the first success regardless of whether the same problem type appeared again.

\subsubsection{Participants continue speaking more after the first success only in the Same condition}



We then tested how speech density changed after the first success. In the Same condition, participants spoke more on post-success trials than on the first-success trial ($\Delta = 0.074$, $p < .0001$). In the Different condition, this change was not reliable ($\Delta = 0.007$, $p = .536$). A phase-by-condition interaction confirmed that the post-success change differed by condition ($\beta = -0.066$, $p < .001$).


We observed the same condition difference when we divided each trial into five
equal-duration segments. In the Same condition, participants spoke more on post-success trials than on first-success trials in bins 1--4 (all $p \le .0016$), but not in bin 5
($p = .737$). In the Different condition, no bin showed a reliable difference
(all $p \ge .240$; Fig.~\ref{fig:speech_density}). Thus, after the first success, participants continued to spend more of the trial speaking only when the same problem type repeated.

\subsubsection{Participants speak faster across phases in the Same condition}


Finally, we analyzed speech rate across phases. In the Same condition, participants spoke faster at the first success than on pre-success trials ($\Delta = 0.099$, $p = .046$), and faster again on post-success trials ($\Delta = 0.106$, $p = .003$). In the Different condition, speech rate did not change reliably across phases (all relevant $p > .43$).

Together, speech-density and speech-rate results show that participants continued to change how much and how quickly they spoke only when the same problem type repeated.

\subsection{Semantic Signatures of Success and Transfer}
We next asked whether participants talked about different things as they moved from unsuccessful trials, to the first success, to later trials. We tested whether trial correctness, condition, and phase could be differentiated from what participants said.

\begin{figure}[t]
  \centering
  \includegraphics[width=\linewidth]{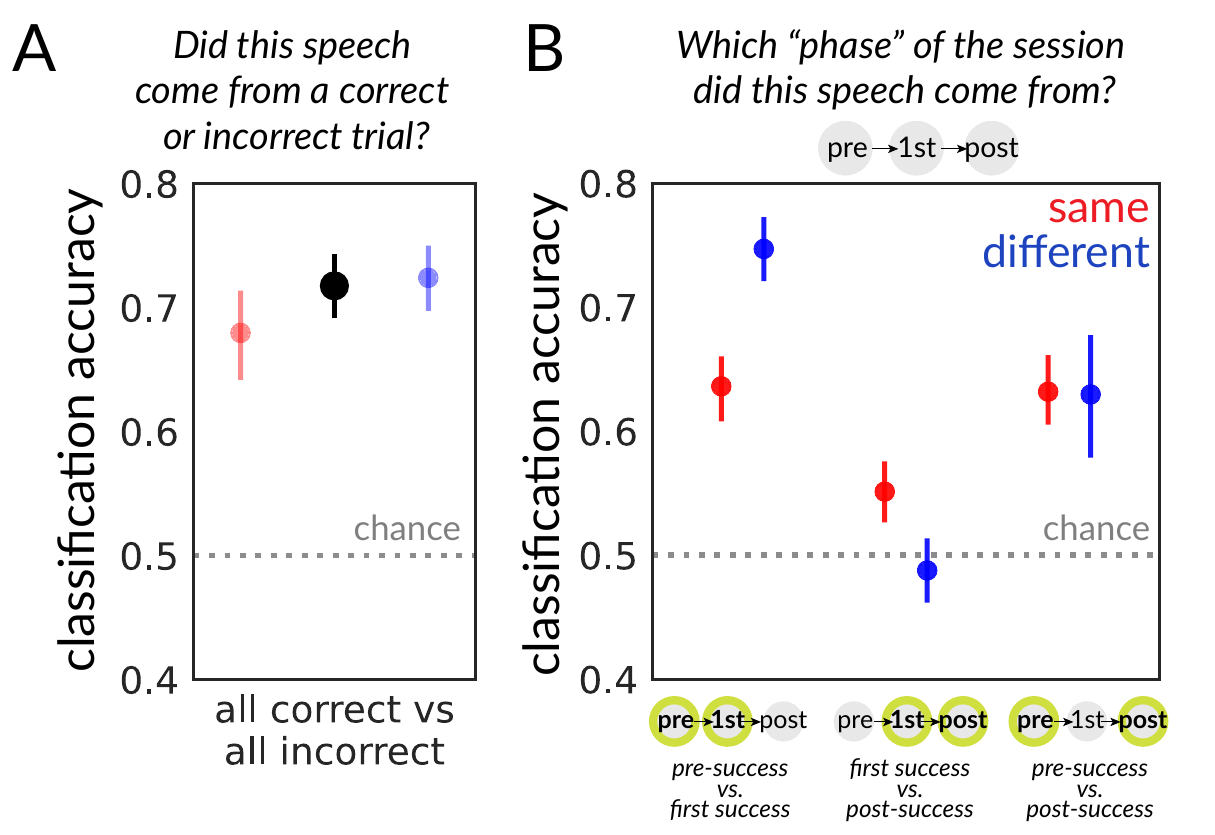}
  \caption{Logistic-regression classification from semantic embeddings of speech utterances. (A) Correct vs.\ incorrect trials by condition (overall in black). (B) Pairwise classification of problem-solving phase (pre, first, post) within each condition. Error bars indicate participant-bootstrap 95\% confidence intervals; dotted lines denote chance.}
  \label{fig:semantics}
\end{figure}

\subsubsection{Participants talk about different things on correct and incorrect trials}



We first tested how easily correct trials could be differentiated from incorrect trials based only on what participants said. Aggregating across the two conditions, correctness was predicted above chance after controlling for task type and trial number (accuracy = .717, 95\% CI [.690, .742]). Classification accuracy was also above chance within each condition (Same: .679, 95\% CI [.643, .714]; Different: .724, 95\% CI [.696, .750]).

We then asked whether participants said different things in the Same and Different conditions when working on the same problem. We focused on Trials~2--5, when Same participants had previously encountered the problem type and Different participants had not. For each problem, we trained classifiers to distinguish condition from utterance embeddings. Classification remained at chance (mean accuracy across problems = .492, 95\% CI [.473, .509]).

Thus, what participants said distinguished correct from incorrect trials, but not Same from Different trials on the same problem.

\subsubsection{What participants say changes across phases in the Same condition}

We next asked how what participants said changed across phases when the same problem type repeated. In the Same condition, we compared utterances from pre-success, first-success, and post-success trials.




Classification accuracy was above chance for these phase contrasts after controlling for task type. Pre-success trials could be distinguished from first-success trials (accuracy = .636, 95\% CI [.608, .660]) and from post-success trials (accuracy = .632, 95\% CI [.605, .661]). First-success trials could also be distinguished from post-success trials, although less accurately (accuracy = .551, 95\% CI [.526, .576]).

Therefore, in the Same condition, what participants said differed between earlier failures and later successes, and continued to change after the first success.

\subsubsection{Participants continue to change what they say only in the Same condition}


We then compared phase-related changes in what participants said across the Same and Different conditions (Fig.~\ref{fig:semantics}B). In each condition, we tested whether classifiers could distinguish pre-success, first-success, and post-success trials.


In both conditions, classifiers distinguished what participants said before the first success from what they said at or after the first success. Pre- vs.\ first-success classification was above chance in both conditions (Same: accuracy = .636, 95\% CI [.608, .660]; Different: accuracy = .747, 95\% CI [.721, .773]). Pre- vs.\ post-success classification was also above chance (Same: .632, 95\% CI [.605, .661]; Different: .630, 95\% CI [.579, .678]).


But the conditions diverged after the first success. Classifiers distinguished
first-success from post-success trials in the Same condition (accuracy = .551,
95\% CI [.526, .576]), but not in the Different condition (accuracy = .488,
95\% CI [.462, .514]).



Taken together, these findings suggest that what participants talked about reliably shifted around the first time they successfully solved a problem in both conditions. After the first success, however, what they said continued to change only when they kept encountering the same kind of problem, not when they kept encountering different kinds of problems.

\subsection{Reasoning-Move Dynamics Across Success and Transfer}


We next asked what role each utterance played in participants' problem solving. Each utterance received one reasoning-move label, such as proposing a candidate move, evaluating a candidate, or categorizing the problem type. To account for differences in how much participants spoke, we analyzed participant-normalized proportions of each label. We focused on two patterns: whether participants began to categorize the problem type, and how many candidate moves they considered across the different trial phases.

\subsubsection{Participants categorize the problem type after the first success only in the Same condition}
We first asked whether participants explicitly named the problem type or noted
that it resembled an earlier problem, which were utterances labeled as
\textbf{Categorization}.



Participants rarely categorized the problem type, but they did so more often after the first success in the Same condition. From first-success to post-success trials, the proportion of categorization utterances increased by 1.8 percentage points ($\Delta = +.018$, 95\% CI [.007, .030]), rising from 0.3\% on first-success trials to 2.1\% on post-success trials (approximately a sevenfold increase). Categorization also increased from pre-success to post-success trials ($\Delta = +.019$, 95\% CI [.004, .038]). In the Different condition, categorization did not increase from first-success to post-success trials ($\Delta = -.002$, 95\% CI [$-.006$, .001]) or from pre-success to
post-success trials ($\Delta = +.001$, 95\% CI [.000, .004]), and remained near zero on post-success trials (mean = .001, 95\% CI [.000, .001]).

Thus, participants did not simply categorize more after any success. They
categorized more only when the same problem type repeated, suggesting that some participants explicitly recognized the discovered structure.

\subsubsection{Participants consider fewer distinct candidate moves after the first success only in the Same condition}

We next asked how participants considered candidate moves across phases. We analyzed two measures: the proportion of utterances labeled as proposals (\emph{proposal proportion}) and the number of distinct candidate moves considered on each trial.


\textbf{Distinct candidate moves:} In both conditions, participants considered
fewer distinct candidate moves at the first success than on pre-success trials
(Same: $\Delta=-2.11$, 95\% CI [$-2.97$, $-1.18$]; Different: $\Delta=-2.24$, 95\% CI [$-3.07$, $-1.41$]). After the first success, the conditions diverged. In the Same condition, participants considered fewer distinct candidate moves on post-success trials than on first-success trials ($\Delta=-0.58$, 95\% CI [$-1.06$, $-0.13$]); in the Different condition, the number of distinct candidate moves did not change ($\Delta=+0.05$, 95\% CI [$-0.65$, $0.77$]).


\textbf{Proposal proportion in speech:} The conditions also differed in proposal proportion. In the Different condition, proposal utterances made up a larger share of speech on first-success trials than on pre-success trials ($\Delta = +0.087$, 95\% CI [0.028, 0.156]) and remained higher on post-success trials than on pre-success trials (pre~$\rightarrow$~post: $\Delta = +0.074$, 95\% CI [0.024, 0.126]). In the Same condition, proposal proportion did not increase from pre-success to first-success trials, but it was higher on post-success trials than on pre-success trials (pre~$\rightarrow$~post: $\Delta = +0.056$, 95\% CI [0.010, 0.102]).


Together, these results suggest that participants in both conditions considered fewer distinct candidate moves at the first success than on pre-success trials. 
After the first success, only participants in the Same condition reduced the number of distinct candidate moves in their speech further. However, participants in both conditions spent more of their reasoning moves generating proposed actions post-success relative to pre-success. 

\section{Discussion}
\begin{figure}[t]
  \centering
  \includegraphics[width=\linewidth]{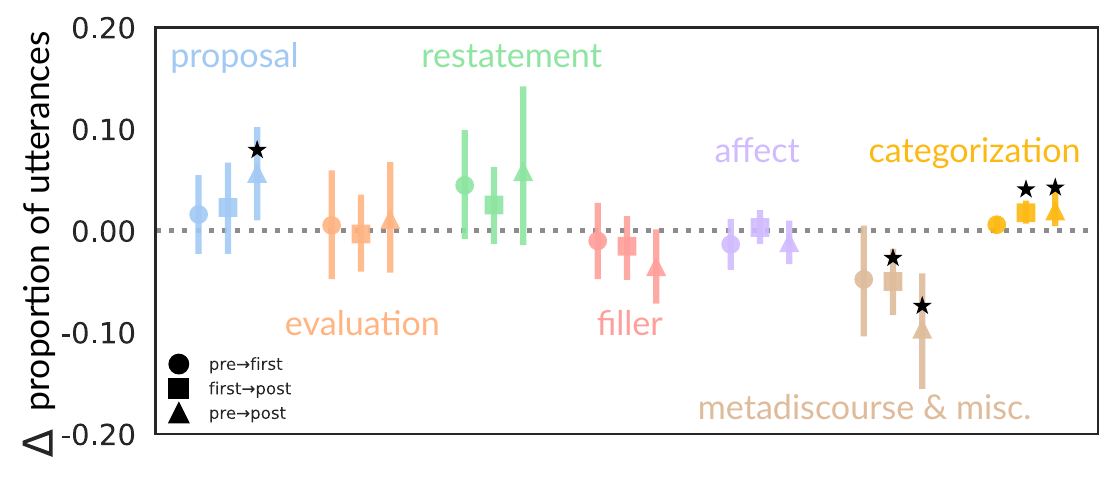}
    \caption{Changes in reasoning move proportions across problem-solving phases in the Same condition, normalized within participants. Each point reflects the change in proportion of a given reasoning category for a specific phase transition (pre~$\rightarrow$~first, first~$\rightarrow$~post, pre~$\rightarrow$~post). Error bars indicate participant-bootstrap 95\% confidence intervals; asterisks mark effects whose confidence intervals exclude zero.}
  \label{fig:reasoning_moves}
\end{figure}

Insight problems require a shift in how the problem is represented \citep{newell1972human, kaplan1990search}.
Although such shifts cannot be observed directly, the first correct solution plausibly marks the moment when one has just occurred.
We therefore used the first success on a matchstick arithmetic problem as our proxy for restructuring, and analyzed think-aloud speech to ask whether this transition was visible and whether participants continued to change how they spoke when the discovered structure could be reused.

First, the first success was marked by changes in speech across multiple levels. 
Participants spoke more on first-success trials, and what they said differed from earlier failures. 
Participants also considered fewer distinct candidate moves at the first success than on earlier trials, consistent with a narrower search.
Thus, the first success did more than change task outcome: it changed how participants reasoned aloud.

Second, participants continued to change how they spoke after the first success only when the discovered structure could be reused.
In the Same condition, participants spoke for a higher fraction of each trial after the first success, sped up, and continued to talk about different things across successful trials.
In the Different condition, participants changed what they talked about around the first success but not on later trials.
These findings suggest that transfer changes how people continue to speak, beyond improvements in performance.

The clearest sign of transfer was that participants began to recognize and name the problem type. 
Only when problems shared a type did participants increasingly produce Categorization statements, explicitly naming the problem type or similarity to prior problems. 
At the same time, participants generated fewer distinct proposal candidates, consistent with a shift from broad exploration toward applying the discovered structure. 
Together, these findings suggest that when the discovered structure can be reused, participants can increasingly report on it, even if the underlying mental operations remain only partially accessible \citep{ericsson1980verbal}.

Several limitations qualify these conclusions. 
We used the first success as our operational marker of a candidate restructuring event, but did not directly measure subjective Aha experiences on each trial. 
Think-aloud protocols also provide an incomplete window into cognition, and automated transcription and labeling can introduce noise. 
Finally, matchstick arithmetic is a constrained insight paradigm, and other domains may show different verbal signatures. 
Future work could combine speech analyses with trial-level ``Aha'' ratings or metacognitive predictions, test whether early verbal features predict later transfer, and examine whether encouraging abstraction causally supports reuse.

More broadly, this work shows that what people say while solving problems can reveal patterns of success and transfer that performance alone misses.

\section{Data and Code Availability}
The preregistration for this experiment can be found \href{https://aspredicted.org/ui5i7u.pdf}{here}. Full code and data are available on \href{https://github.com/cogtoolslab/speech-insight-transfer}{GitHub}.

\section{Acknowledgments}
We thank Jay McClelland, Ben Prystawski, Daniel Wurgaft, and Noah Goodman for helpful feedback on this work. We also thank Jeffrey Mu for assistance with reasoning-move annotations, including helping to develop the annotation methodology and label the pilot data. J.E.F. is supported by NSF CAREER Award \#2436199, NSF DRL \#2400471, and awards from the Stanford Human-Centered AI Institute (HAI) and Stanford Accelerator for Learning.


\printbibliography

\end{document}